\documentclass[conference]{IEEEtran}
\IEEEoverridecommandlockouts
\usepackage{cite}
\usepackage{amsmath,amssymb,amsfonts}
\usepackage{algorithmic}
\usepackage{graphicx}
\usepackage{textcomp}
\usepackage{xcolor}
\def\BibTeX{{\rm B\kern-.05em{\sc i\kern-.025em b}\kern-.08em
    T\kern-.1667em\lower.7ex\hbox{E}\kern-.125emX}}
\begin{document}

\title{Text me the data: Generating Ground Pressure Sequence from Textual Descriptions for HAR\\
\thanks{The research was supported by the BMBF (German Federal Ministry of Education and Research) in the project VidGenSense (01IW21003).}
}

\author{\IEEEauthorblockN{Lala Shakti Swarup Ray}
\IEEEauthorblockA{\textit{DFKI} \\
Kaiserslautern, Germany \\
lala\_shakti\_swatup.ray@dfki.de}
\and
\IEEEauthorblockN{Bo Zhou}
\IEEEauthorblockA{\textit{RPTU Kaiserslautern-Landau \& DFKI} \\
Kaiserslautern, Germany \\
bo.zhou@dfki.de}
\and
\IEEEauthorblockN{Sungho Suh}
\IEEEauthorblockA{\textit{RPTU Kaiserslautern-Landau \& DFKI} \\
Kaiserslautern, Germany \\
sungho.shu@dfki.de}
\and
\IEEEauthorblockN{Lars Krupp}
\IEEEauthorblockA{\textit{RPTU Kaiserslautern-Landau \& DFKI} \\
Kaiserslautern, Germany \\
lars.krupp@dfki.de}
\and
\IEEEauthorblockN{Vitor Fortes Rey}
\IEEEauthorblockA{\textit{RPTU Kaiserslautern-Landau \& DFKI} \\
Kaiserslautern, Germany \\
vitor\_fortes.rey@dfki.de}
\and
\IEEEauthorblockN{Paul Lukowicz}
\IEEEauthorblockA{\textit{RPTU Kaiserslautern-Landau \& DFKI} \\
Kaiserslautern, Germany \\
paul.lukowicz@dfki.de}
}

\maketitle

\begin{abstract}
In human activity recognition (HAR), the availability of substantial ground truth is necessary for training efficient models. However, acquiring ground pressure data through physical sensors itself can be cost-prohibitive, time-consuming. To address this critical need, we introduce Text-to-Pressure (T2P), a framework designed to generate extensive ground pressure sequences from textual descriptions of human activities using deep learning techniques. We show that the combination of vector quantization of sensor data along with simple text conditioned auto regressive strategy allows us to obtain high-quality generated pressure sequences from textual descriptions with the help of discrete latent correlation between text and pressure maps. 
We achieved comparable performance on the consistency between text and generated motion with an R squared value of 0.722, Masked R squared value of 0.892, and FID score of 1.83. Additionally, we trained a HAR model with the the synthesized data and evaluated it on pressure dynamics collected by a real pressure sensor which is on par with a model trained on only real data. Combining both real and synthesized training data increases the overall macro F1 score by 5.9 percent. 
\end{abstract}


\section{Introduction}
Ground pressure data collected from humans plays an indispensable role in a wide array of applications, spanning from Human Activity Recognition (HAR) \cite{hou2022crack} to gait analysis \cite{ali2023graphene}, healthcare monitoring \cite{meng2022wearable}, rehabilitation programs \cite{han2022smart}, robotics development \cite{fu2022fingerprint}, and human-robot interaction \cite{fan2022enabling}. 
However, a significant challenge arises when it comes to collecting ground pressure data, as it traditionally relies on physical sensors such as pressure pads, smart mats, and pressure insoles. These conventional sensing devices present inherent limitations related to portability, versatility and durability that makes the data collection process both time-consuming and resource-intensive. This obstacle becomes particularly daunting when striving to amass a diverse and substantial dataset encompassing various users and a multitude of activity variations.

To address this challenge, researchers have explored innovative approaches to synthesize pressure data from readily available modalities like RGB images \cite{ray2023pressim} and depth images \cite{clever2022bodypressure} but those models need real data from another modality as input making the whole purpose redundant. Another intriguing approach involves using activity labels themselves to generate pressure sensor data. Recent advancements in generative models, especially within the realm of transformers, have revolutionized the field of data generation from textual descriptions. Large Language Models (LLMs) like CLIP \cite{radford2021learning} and the GPT series \cite{floridi2020gpt} have emerged as game-changers. These LLMs, equipped with billions of parameters, exhibit remarkable capabilities in comprehending and generating text. Importantly, they can transform the text into various data formats, including images \cite{rombach2022high} and 3D pose data \cite{kim2023flame}. This transformative technology empowers researchers and practitioners by enabling the creation of diverse and high-quality datasets directly from text, expanding the boundaries of what AI can accomplish.
In the context of ground pressure sensors, LLMs offer the potential to convert textual descriptions of human activities into synthetic ground pressure data. This approach not only addresses the challenges of slow and costly data collection but also serves as valuable training material for models like T2P. By adopting this method, T2P can learn to generate highly accurate ground pressure sequences from text, significantly enhancing their effectiveness in HAR tasks. 

To summarize, this paper makes the following contributions:

\begin{itemize}
\item A method of sensor data synthesis from texts that enables training a sensor-based HAR system (dynamic planar pressure) that outperforms a model trained with real data, validated with a real sensor system.
\item A large synthetic dataset consisting of 240 activity descriptions, with 720 unique pressure map sequences of 86,400 frames and a real dataset with eight activity descriptions from ten individuals, totaled 16,256 pressure frames. 
\end{itemize}

\section{Approach}
\label{sec:app}
\subsection{Dataset Synthesis Pipeline}
\begin{figure}
\begin{center}
\includegraphics[width=1\linewidth]{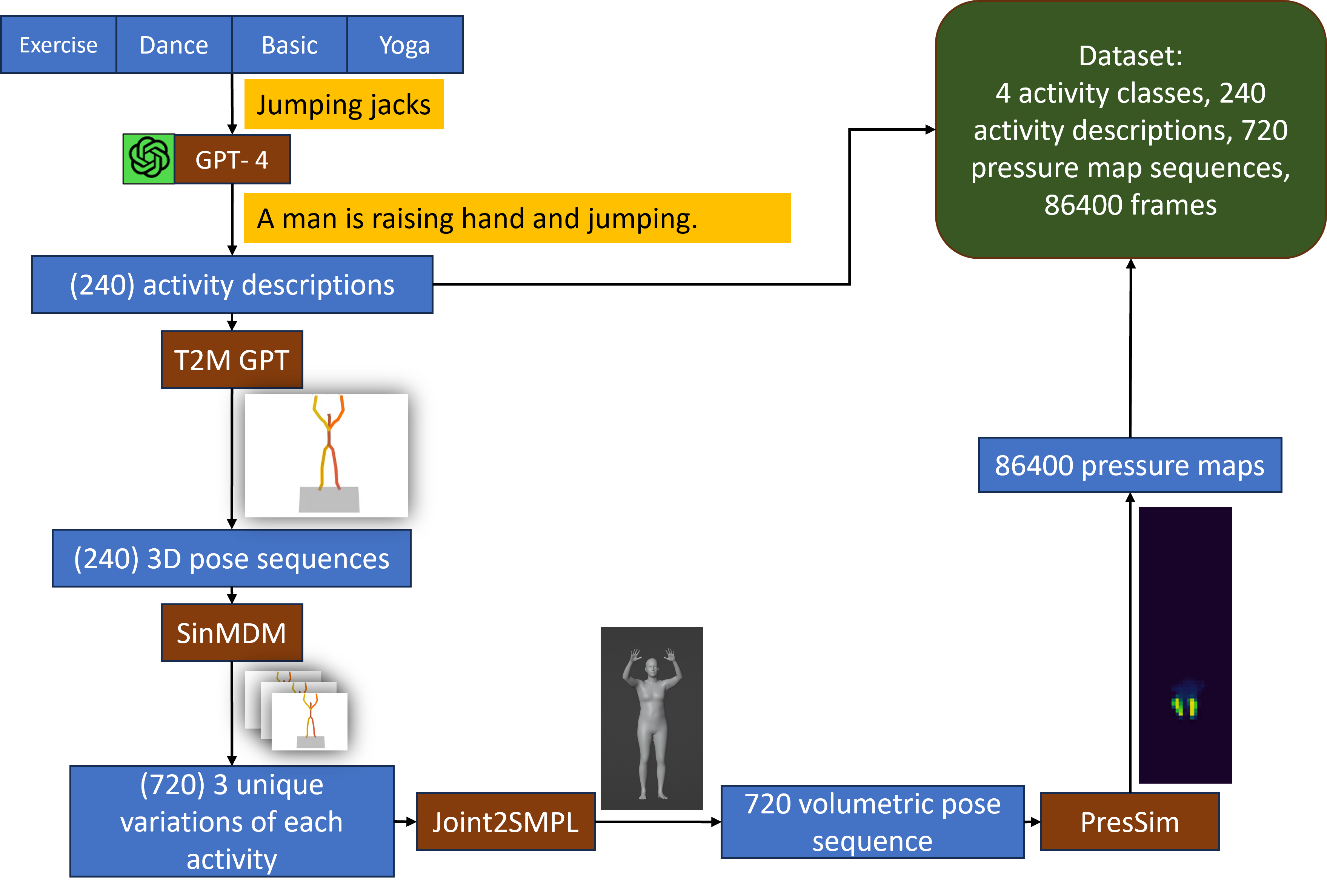}
\end{center}
   \caption{Data generation pipeline that combines GPT-4, T2M-GPT, SinMDM, joint2SMPL, and PresSim to generate pressure map sequences from activity description for training T2P.}
\label{fig:2}
\end{figure}

To generate the extensive dataset encompassing four distinct activity categories – basic, dance, yoga, and workout, all revolving around lower body movements and semi-static actions, a meticulous and intricate procedure was employed. This comprehensive process hinged on the utilization of state of the art technologies, including GPT-4 \cite{openai2023gpt4}, T2M-GPT \cite{zhang2023t2m}, SinMDM \cite{raab2023single}, joint2smpl \cite{zuo2021sparsefusion}, and PresSim \cite{ray2023pressim}, as illustrated in Figure \ref{fig:2}.

The primary rationale behind opting for synthesized motion as opposed to utilizing pre-existing motion datasets such as HumanML3D \cite{Guo_2022_CVPR} or the KIT-ML dataset \cite{Plappert2016} is rooted in the unique nature of the target pressure data we aimed to generate. This data pertains to an 80x28 array of pressure-sensing mats, which permits only limited movement across its surface. Straying from the defined surface wouldn't capture the desired pressure data accurately.

The initial step involved the manual collection of sixty distinct motion types associated with each activity class. Subsequently, we employed prompt engineering with GPT-4 to deconstruct these activities into their elemental motions, stripping away any external environmental context. This was achieved by assigning "role" and "content" and providing T2M-GPT examples as reference prompts. For instance, an activity like \textit{"Push-ups"} would be transformed into an activity description like \textit{"A person assumed an all-fours position, facing downwards, then lowered their upper body toward the ground and lifted it back up using only their arms."}

Following this, T2M-GPT\cite{zhang2023t2m} was employed to translate these activity descriptions into 3D kinematic pose sequences. Each pose sequence comprised 22 joints and 120 frames for each motion. To maintain the quality of the generated motions, they underwent manual inspection to identify and eliminate any inaccuracies or errors.

To introduce diversity and variability into the dataset, SinMDM \cite{raab2023single} is harnessed to generate three unique motion sequences for each activity, with a random number of frames, resulting in a total of 720 3D pose sequences spanning 86400 frames. This approach significantly enriched the dataset by introducing complexity and diversity.

To prepare these generated 3D poses for use with PresSim, they needed to be converted into volumetric poses in SMPL format \cite{loper2023smpl}. To achieve this, we employed the joint2smpl technique \cite{zuo2021sparsefusion}, which expertly converted the 3D pose sequences into the desired SMPL format.Finally, PresSim was utilized to transform these volumetric poses into 720 distinct pressure sequences. This comprehensive process yielded a dataset that included 240 unique activity descriptions and a total of 86,400 frames.
\subsection{Validation Data Collection}
We also collected a real dataset to validate our model which is explained in detail in Section \ref{sec:res}.
We used a Fitness-mat from Sensing.Tex \cite{sensingtex_2023} with a \(80 \times 28\) sensor grid having a sensing area of \(560\times 1680\) \(mm \) with separate sensor area of \(12\times 16 \) \(mm \) with measuring range of 0-5000 millimeters of mercury (mmHg). 
The hardware consists of a non-elastic, non-slippery yoga mat made with Thermoplastic Polyolefins(TPE) polymers having a range of operation from 5\% – 70\% humidity and 15$ ^\circ C $  – 45 $ ^\circ C $.
The detailed data statistics are provided in Table \ref{tab:data}.

\begin{table}[!t]
\centering
\caption{ Data statistics including subject mass (\(kilogram\)), height (\(centimeter\)) and gender.}
\label{tab:data}
\begin{tabular}{|l|c|c|c|c|}
\hline
\textbf{Subject} & \textbf{Mass (\(kg\))} & \textbf{Height (\(cm\))} & \textbf{Gender} & \textbf{Frames} \\
\hline
1 & 57.7 & 162 & Female & 802 \\
2 & 94.7 & 172 & Male & 1247 \\
3 & 76.3 & 183 & Male & 1045 \\
4 & 74.3 & 175 & Male & 1031 \\
5 & 53.3 & 153 & Female & 743 \\
6 & 62.3 & 173 & Male & 865 \\
7 & 77.2 & 184 & Male & 1069 \\
8 & 72.2 & 178 & Male & 989 \\
9 & 60.2 & 156 & Male & 839 \\
10 & 67.3 & 178 & Female & 932 \\
Total & - & - & - & 16256 \\
\hline
\end{tabular}
\end{table}

\subsection{Architecture}
\begin{figure}
\begin{center}
\includegraphics[width=1\linewidth]{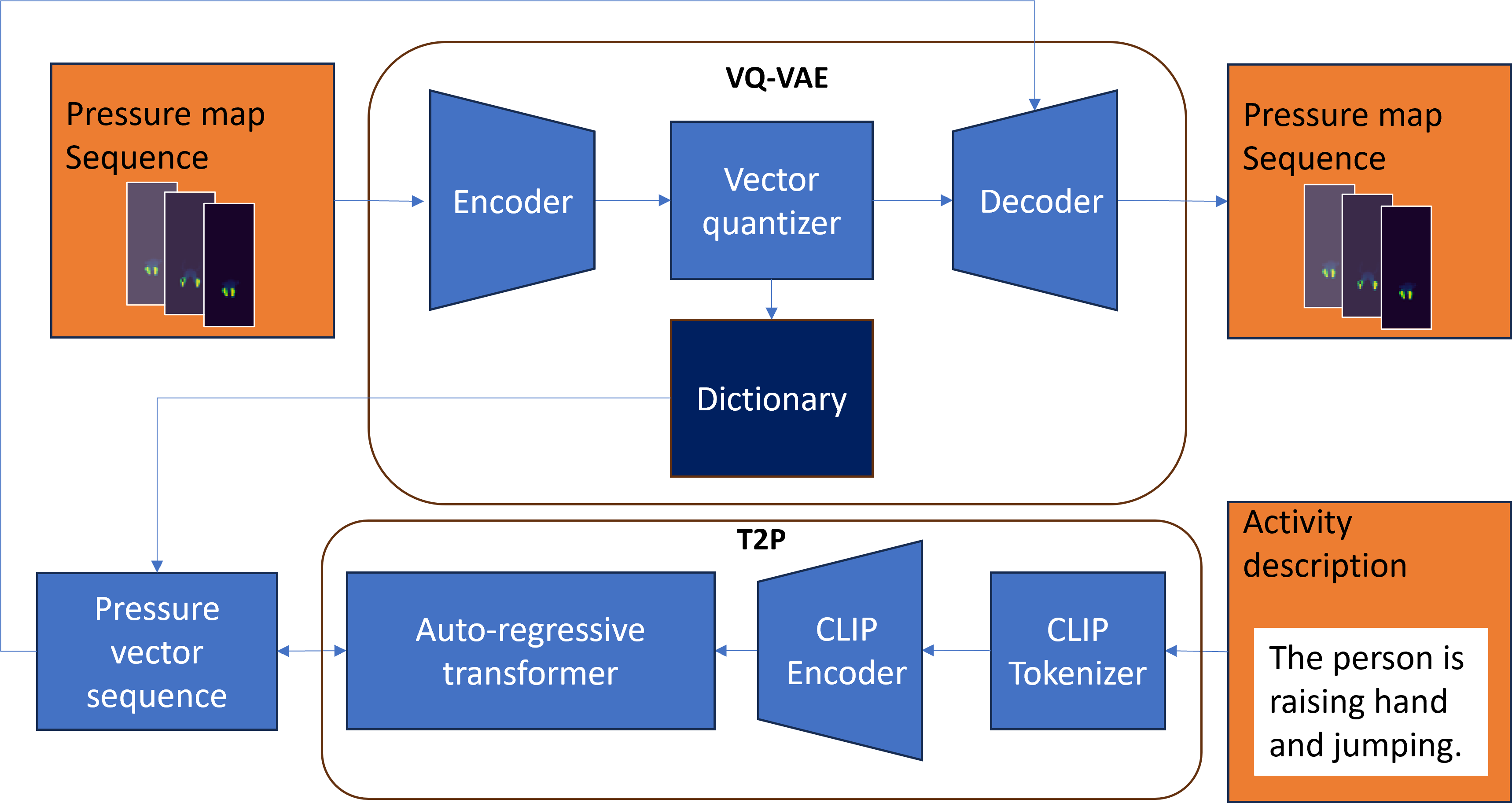}
\end{center}
   \caption{Illustration of the T2P architecture that utilizes Variational auto-encoder (VQ-VAE) and auto-regressive transformer (T2P) module to train the model.}
\label{fig:1}
\end{figure}
\paragraph{Vector Quantized Variational Autoencoder (VQ-VAE)}
Inspired by T2M-GPT, we discretized the Pressure map sequence by employing a VQ-VAE given in Figure \ref{fig:1} to create a dictionary for the pressure sequences. This process facilitates the discrete representation of continuous pressure data, enhancing its suitability for modeling. The resulting dictionary serves as a compact and structured representation of intricate pressure information. 
The autoencoder follows a ResNet-inspired simple 1D CNN blocks architecture with residual blocks (RB) in both encoder and decoder, maintaining the same input and output size for the Pressure map sequence. ReLu activation is utilized in the network, and a vector quantizer is inserted between the encoder and decoder to convert extracted features into vector sequences. Training involves a combination of reconstruction loss (\(L_{\text{r}}\)) and quantization loss (\(L_{\text{q}}\)).
To enhance the robustness of our dictionary, scheduled annealing is implemented. Scheduled annealing dynamically adjusts the weights for reconstruction (\(w_{\text{r}}\)) and quantization (\(w_{\text{q}}\)) throughout training. The weights are defined as functions of training step (\(t\)) and contribute to the total loss (\(L(t)\)), balancing reconstruction and quantization losses.
\[ L(t) = w_{\text{r}}(t) \cdot L_{\text{r}}(t) + w_{\text{q}}(t) \cdot L_{\text{q}}(t) \]
Additionally, we employ exponential moving average (EMA) to stabilize training and enhance the quality of learned representations in the dictionary. EMA prevents rapid changes in the dictionary, ensuring stability and meaningful feature capture from the input data, thereby contributing to improved training and generative capabilities.
\[ \text{EMA}_t = (1 - \alpha) \cdot \text{EMA}_{t-1} + \alpha \cdot \text{Dictionary}_t \]
Here, \(\text{EMA}_t\) represents the Exponential Moving Average at time \(t\), \(\alpha\) is the smoothing factor for EMA (computed as \(\alpha = \frac{2}{N + 1}\) with \(N\) as a hyper-parameter), and \(\text{Dictionary}_t\) denotes the dictionary entries at training step \(t\).

\paragraph{T2P}
In T2P, we harnessed the power of a pre-trained CLIP encoder to derive meaningful text embeddings from natural language text, denoted as \(E_{\text{CLIP}}(T)\), forming the foundation for our subsequent modeling efforts.
The pressure sequences, denoted as \(P = [p_1, p_2, \ldots, p_T]\), were then transformed into a sequence of indices, \(P' = [p'_1, p'_2, \ldots, p'_{T/l}, \text{End}]\), where \(p'_i \in [1, 2, \ldots, p'_{T/l}]\). Here, \(T/l\) indicates the division of the original sequence into segments, each of length \(l\), with indices \(p'_i\) drawn from a pre-learned dictionary.
Mathematically, this conversion process is represented as:
\[P \rightarrow P' = [p'_1, p'_2, \ldots, p'_{T/l}, \text{End}]\]
The obtained indices \(p'_i\), coupled with the text conditioning \(E_{\text{CLIP}}(T)\), serve as crucial inputs to our auto-regressive transformer to generate a sequence of vectors. At each step, the model predicts the next index based on prior predictions and the text conditioning, with the "End" token signaling the sequence termination.
Mathematically, the auto-regressive sequence generation process for synthesized pressure vectors \(P'' = [p''_1, p''_2, \ldots, \text{End}]\), given text conditioning \(E_{\text{CLIP}}(T)\), is expressed as:
\[E_{\text{transformer}}(P', E_{\text{CLIP}}(T)) \rightarrow [p''_1, p''_2, \ldots, \text{End}]\]
The resulting vector sequences are then transformed back into pressure sequences using the pre-trained VQ-VAE decoder.

\section{Experimental Results}
\label{sec:res}
\paragraph{Implementation details}
The VQ-VAE model trained on all 86,400 pressure frames using 10-fold cross-validation. Early stopping and a decaying learning rate scheduler (\(1e^{-3}\)) were employed to prevent overfitting.
For T2M, the dataset was divided into 66,200 training sets, 5,120 test sets, and 15,080 validation sets. Both models, implemented in PyTorch 2.0, were trained on a cluster with 128GB memory, four processors, and two Nvidia RTX A6000 graphics cards.
\paragraph{Qualitative analysis}
\begin{figure}
\begin{center}
\includegraphics[width=1\linewidth]{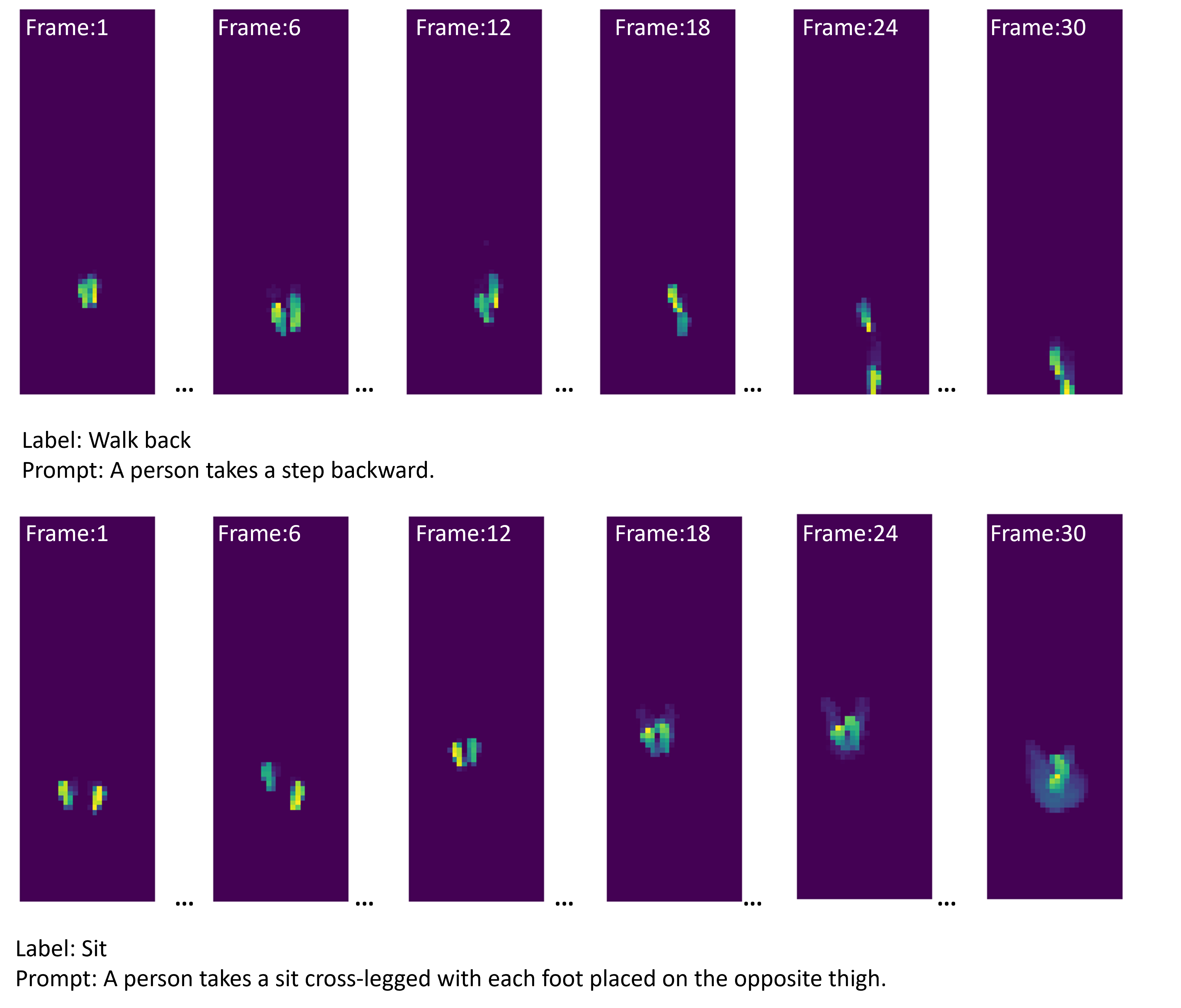}
\end{center}
   \caption{visualization of T2P results for two activities walk back and sit.}
\label{fig:6}
\end{figure}
In Figure \ref{fig:6}, we showcase generated Pressure sequences paired with corresponding text prompts. GPT-4 prompt engineering is employed to convert labels into text prompts, which are then used as input for T2P to generate the pressure sequence. While acknowledging that our method may not achieve a perfect representation of the underlying motion details, it effectively communicates the original activity label through the resulting pressure sequences.
\paragraph{Quantitative analysis}
We used FID metrics that measures the similarity between real and generated pressure maps, and binarized R2 that assesses the accurateness of point of contact in our generated data.In our ablation studies, the VQ-VAE was enhanced with residual blocks (RB), scheduled annealing (SA), and exponential moving averages (EMA), resulting in a notable 1.802 improvement in the FID score \ref{tab:vqvae_ablation}. Comparisons were made with a baseline T2P model that uses extracted features instead of quantized vectors to train the autorgressive transformer, showing the latter's effectiveness despite the complexity of pressure data \ref{tab:comparison}. Our model was also compared with PressureTransferNet, another model to generate synthetic pressure sequence from existing pressure using style transfer techniques, revealing a 0.102 improvement in binarized $R^2$ score without the need for physical pressure data. 
Evaluation of T2P using real pressure data from an 80x28 pressure sensing mattress involved creating synthetic pressure sequences for diverse activity labels. The HAR model, trained on a combination of synthetic and real data, outperformed models trained solely on real data, highlighting the significance of data diversity and augmentation. This approach, leveraging synthetic data for controlled exploration and enriching training datasets, enhances model accuracy and adaptability to real-world pressure data complexities and variations \ref{tab:har}. 
\begin{table}[!t]
\centering
\caption{Ablation Studies on VQ-VAE using RB, SA and EMA.}
\label{tab:vqvae_ablation}
\begin{tabular}{|l|c|c|}
\hline
\textbf{Model} & \textbf{Dictionary size } & \textbf{FID} \\
\hline
VQ-VAE & 128 & 3.542 \\
VQ-VAE & 256 & 2.012 \\
VQ-VAE & 512 & \textbf{1.813} \\
\hline
VQ-VAE + RB & 512 & 0.767 \\
VQ-VAE + RB + SA & 512 & 0.336 \\
VQ-VAE + RB + SA + EMA & 512 & \textbf{0.011} \\
\hline
\end{tabular}
\end{table}

\begin{table}[!t]
\centering
\caption{Baseline vs. VQ-VAE T2P using $R^2$, Binarized $R^2$, and FID}
\label{tab:comparison}
\begin{tabular}{|l|c|c|c|}
\hline
\textbf{Training approach} & \textbf{Binarized $R^2$} & \textbf{$R^2$} & \textbf{FID} \\
\hline
Baseline T2P & 0.635 & 0.574 & 7.311 \\
VQ-VAE T2P & \textbf{0.892} & \textbf{0.722} & \textbf{1.830} \\
\hline
\end{tabular}
\end{table}

\begin{table}[!t]
\centering
\caption{F1 score of SHAR model trained on synthetic data vs. real data.}
\label{tab:har}
\begin{tabular}{|l|c|}
\hline
\textbf{HAR training set} & \textbf{macro F1-score}\\
\hline
Synthetic (T2P) generated data & 0.502 $\pm$ 0.031\\
Real pressure data & 0.742 $\pm$ 0.038\\
Synthetic (PressureTransferNET + T2P) data & 0.737 $\pm$ 0.035\\
Real+ synthetic (PressureTransferNET + T2P) data & \textbf{0.801 $\pm$ 0.027}\\
\hline
\end{tabular}
\end{table}

\section{Conclusion}
\label{sec:con}
In this work, we have introduced T2P, an innovative framework that combines vector quantization of sensor data with a simple auto-regressive method, to generate high-quality ground pressure sequences that exhibit a discrete latent correlation between text and pressure maps.
Our method demonstrates remarkable results, with $R^2$ of 0.722, binarized $R^2$ of 0.892, and FID of 1.83, indicating strong consistency between the text and the generated pressure. Furthermore, we leveraged the synthesized data to train a HAR model, which increased the overall macro F1 score by approximately 5.9 percent when trained with a combination of both synthetic and real data.
These findings underscore the potential of this framework to enhance the robustness of HAR models in the pressure sensor domain, showcasing the effectiveness of leveraging current large language models to generate synthetic data for practical applications. Moreover, the versatility of this workflow extends beyond ground pressure data, as it can be applied to other sensor modalities, offering a promising avenue for generating sensor data from textual descriptions across various domains.
\bibliographystyle{IEEEtran}
\bibliography{refs}

\begin{thebibliography}{10}
\providecommand{\url}[1]{#1}
\csname url@samestyle\endcsname
\providecommand{\newblock}{\relax}
\providecommand{\bibinfo}[2]{#2}
\providecommand{\BIBentrySTDinterwordspacing}{\spaceskip=0pt\relax}
\providecommand{\BIBentryALTinterwordstretchfactor}{4}
\providecommand{\BIBentryALTinterwordspacing}{\spaceskip=\fontdimen2\font plus
\BIBentryALTinterwordstretchfactor\fontdimen3\font minus \fontdimen4\font\relax}
\providecommand{\BIBforeignlanguage}[2]{{%
\expandafter\ifx\csname l@#1\endcsname\relax
\typeout{** WARNING: IEEEtran.bst: No hyphenation pattern has been}%
\typeout{** loaded for the language `#1'. Using the pattern for}%
\typeout{** the default language instead.}%
\else
\language=\csname l@#1\endcsname
\fi
#2}}
\providecommand{\BIBdecl}{\relax}
\BIBdecl

\bibitem{hou2022crack}
Y.~Hou, L.~Wang, R.~Sun, Y.~Zhang, M.~Gu, Y.~Zhu, Y.~Tong, X.~Liu, Z.~Wang, J.~Xia \emph{et~al.}, ``Crack-across-pore enabled high-performance flexible pressure sensors for deep neural network enhanced sensing and human action recognition,'' \emph{ACS nano}, vol.~16, no.~5, 2022.

\bibitem{ali2023graphene}
B.~Ali, N.~Faramarzi, U.~Farooq, H.~C. Bidsorkhi, A.~G. D'Aloia, A.~Tamburrano, and M.~S. Sarto, ``Graphene-based smart insole sensor for pedobarometry and gait analysis,'' \emph{IEEE Sensors Letters}, 2023.

\bibitem{meng2022wearable}
K.~Meng, X.~Xiao, W.~Wei, G.~Chen, A.~Nashalian, S.~Shen, X.~Xiao, and J.~Chen, ``Wearable pressure sensors for pulse wave monitoring,'' \emph{Advanced Materials}, vol.~34, no.~21, p. 2109357, 2022.

\bibitem{han2022smart}
Y.~Han, A.~Varadarajan, T.~Kim, G.~Zheng, K.~Kitani, A.~Kelliher, T.~Rikakis, and Y.-L. Park, ``Smart skin: Vision-based soft pressure sensing system for in-home hand rehabilitation,'' \emph{Soft Robotics}, vol.~9, no.~3, pp. 473--485, 2022.

\bibitem{fu2022fingerprint}
X.~Fu, J.~Dong, L.~Li, L.~Zhang, J.~Zhang, L.~Yu, Q.~Lin, J.~Zhang, C.~Jiang, J.~Zhang \emph{et~al.}, ``Fingerprint-inspired dual-mode pressure sensor for robotic static and dynamic perception,'' \emph{Nano Energy}, vol. 103, p. 107788, 2022.

\bibitem{fan2022enabling}
X.~Fan, D.~Lee, L.~Jackel, R.~Howard, D.~Lee, and V.~Isler, ``Enabling low-cost full surface tactile skin for human robot interaction,'' \emph{IEEE Robotics and Automation Letters}, vol.~7, no.~2, pp. 1800--1807, 2022.

\bibitem{ray2023pressim}
L.~S.~S. Ray, B.~Zhou, S.~Suh, and P.~Lukowicz, ``Pressim: An end-to-end framework for dynamic ground pressure profile generation from monocular videos using physics-based 3d simulation,'' in \emph{2023 IEEE International Conference on Pervasive Computing and Communications Workshops and other Affiliated Events (PerCom Workshops)}.\hskip 1em plus 0.5em minus 0.4em\relax IEEE, 2023, pp. 484--489.

\bibitem{clever2022bodypressure}
H.~M. Clever, P.~L. Grady, G.~Turk, and C.~C. Kemp, ``Bodypressure-inferring body pose and contact pressure from a depth image,'' \emph{IEEE Transactions on Pattern Analysis and Machine Intelligence}, vol.~45, no.~1, pp. 137--153, 2022.

\bibitem{radford2021learning}
A.~Radford, J.~W. Kim, C.~Hallacy, A.~Ramesh, G.~Goh, S.~Agarwal, G.~Sastry, A.~Askell, P.~Mishkin, J.~Clark \emph{et~al.}, ``Learning transferable visual models from natural language supervision,'' in \emph{International conference on machine learning}.\hskip 1em plus 0.5em minus 0.4em\relax PMLR, 2021, pp. 8748--8763.

\bibitem{floridi2020gpt}
L.~Floridi and M.~Chiriatti, ``Gpt-3: Its nature, scope, limits, and consequences,'' \emph{Minds and Machines}, vol.~30, pp. 681--694, 2020.

\bibitem{rombach2022high}
R.~Rombach, A.~Blattmann, D.~Lorenz, P.~Esser, and B.~Ommer, ``High-resolution image synthesis with latent diffusion models,'' in \emph{IEEE/CVF conference on computer vision and pattern recognition}, 2022.

\bibitem{kim2023flame}
J.~Kim, J.~Kim, and S.~Choi, ``Flame: Free-form language-based motion synthesis \& editing,'' in \emph{Proceedings of the AAAI Conference on Artificial Intelligence}, vol.~37, no.~7, 2023, pp. 8255--8263.

\bibitem{openai2023gpt4}
OpenAI, ``Gpt-4 technical report,'' 2023.

\bibitem{zhang2023t2m}
J.~Zhang, Y.~Zhang, X.~Cun, S.~Huang, Y.~Zhang, H.~Zhao, H.~Lu, and X.~Shen, ``T2m-gpt: Generating human motion from textual descriptions with discrete representations,'' \emph{arXiv preprint arXiv:2301.06052}, 2023.

\bibitem{raab2023single}
S.~Raab, I.~Leibovitch, G.~Tevet, M.~Arar, A.~H. Bermano, and D.~Cohen-Or, ``Single motion diffusion,'' \emph{arXiv preprint arXiv:2302.05905}, 2023.

\bibitem{zuo2021sparsefusion}
X.~Zuo, S.~Wang, J.~Zheng, W.~Yu, M.~Gong, R.~Yang, and L.~Cheng, ``Sparsefusion: Dynamic human avatar modeling from sparse rgbd images,'' \emph{IEEE Transactions on Multimedia}, vol.~23, pp. 1617--1629, 2021.

\bibitem{Guo_2022_CVPR}
C.~Guo, S.~Zou, X.~Zuo, S.~Wang, W.~Ji, X.~Li, and L.~Cheng, ``Generating diverse and natural 3d human motions from text,'' in \emph{Proceedings of the IEEE/CVF Conference on Computer Vision and Pattern Recognition (CVPR)}, June 2022, pp. 5152--5161.

\bibitem{Plappert2016}
\BIBentryALTinterwordspacing
M.~Plappert, C.~Mandery, and T.~Asfour, ``The {KIT} motion-language dataset,'' \emph{Big Data}, vol.~4, no.~4, pp. 236--252, dec 2016. [Online]. Available: \url{http://dx.doi.org/10.1089/big.2016.0028}
\BIBentrySTDinterwordspacing

\bibitem{loper2023smpl}
M.~Loper, N.~Mahmood, J.~Romero, G.~Pons-Moll, and M.~J. Black, ``Smpl: A skinned multi-person linear model,'' in \emph{Seminal Graphics Papers: Pushing the Boundaries, Volume 2}, 2023, pp. 851--866.

\bibitem{sensingtex_2023}
\BIBentryALTinterwordspacing
S.~T. Team, ``Stretchable electronics, smarter connectivity,'' 2023. [Online]. Available: \url{http://sensingtex.com/}
\BIBentrySTDinterwordspacing

\end{thebibliography}

\end{document}